\documentclass[runningheads]{llncs}
\pdfoutput=1
\usepackage{eccv}

\usepackage{eccvabbrv}

\usepackage{graphicx}
\usepackage{booktabs}

\usepackage[accsupp]{axessibility}  % Improves PDF readability for those with disabilities.

\usepackage{floatrow}
\newfloatcommand{capbtabbox}{table}[][\FBwidth]
\usepackage{times}
\usepackage{epsfig}
\usepackage{amsmath}
\usepackage{amssymb}
\usepackage{graphicx}
\usepackage{subcaption}
\usepackage{multirow}
\usepackage{multicol}
\usepackage{floatrow}
\usepackage{algorithm}
\usepackage{algorithmic}
\usepackage{booktabs}
\usepackage{mathrsfs}
\usepackage{colortbl}  %彩色表格需要加载的宏包
\usepackage{array}   %对表列和表格线的设置需要用到array宏包
\usepackage{pifont}
\definecolor{gray0}{gray}{0.9}

\newcommand{\Ours}{\textup{ControlCap}\xspace}

\renewcommand{\Sigma}{\mathfrak{S}}

\def\eqref#1{equation~\ref{#1}}

\def\1{\bm{1}}

\DeclareMathAlphabet{\mathsfit}{\encodingdefault}{\sfdefault}{m}{sl}
\SetMathAlphabet{\mathsfit}{bold}{\encodingdefault}{\sfdefault}{bx}{n}

\usepackage[pagebackref,breaklinks,colorlinks,citecolor=eccvblue]{hyperref}
\usepackage{orcidlink}

\begin{document}

% \title{Controllable Referring Captioner with Multimodal Embedding Bridging} 
% \title{ControlCap: Controllable Referring Expression Generation} 
\title{ControlCap: Controllable Region-level Captioning}

\titlerunning{Controllable Region-level Captioning}


\author{Yuzhong Zhao\inst{1} \and
Yue Liu\inst{1} \and
Zonghao Guo\inst{1}\and
Weijia Wu\inst{2}\and
Chen Gong\inst{3}\and
Fang Wan\inst{1}\thanks{Corresponding Author.}\and
Qixiang Ye\inst{1}}

% % TODO FINAL: Replace with an abbreviated list of authors.
\authorrunning{Y.~Zhao et al.}

% % TODO FINAL: Replace with your institution list.
\institute{University of Chinese Academy of Sciences\and
Zhejiang University\and
University of Virginia}

\maketitle

\begin{abstract}
Region-level captioning is challenged by the caption degeneration issue, which refers to that pre-trained multimodal models tend to predict the most frequent captions but miss the less frequent ones.
In this study, we propose a controllable region-level captioning (ControlCap) approach, which introduces control words to a multimodal model to address the caption degeneration issue. 
In specific, ControlCap leverages a discriminative module to generate control words within the caption space to partition it to multiple sub-spaces.
The multimodal model is constrained to generate captions within a few sub-spaces containing the control words, which increases the opportunity of hitting less frequent captions, alleviating the caption degeneration issue.
Furthermore, interactive control words can be given by either a human or an expert model, which enables captioning beyond the training caption space, enhancing the model's generalization ability.
Extensive experiments on Visual Genome and RefCOCOg datasets show that ControlCap respectively improves the CIDEr score by 21.6 and 2.2, outperforming the state-of-the-arts by significant margins.
Code is available at \url{https://github.com/callsys/ControlCap}.
\end{abstract}
%%%%%%%%摘要修改的很棒

\begin{figure*}[t]
	\includegraphics[width=1\linewidth]{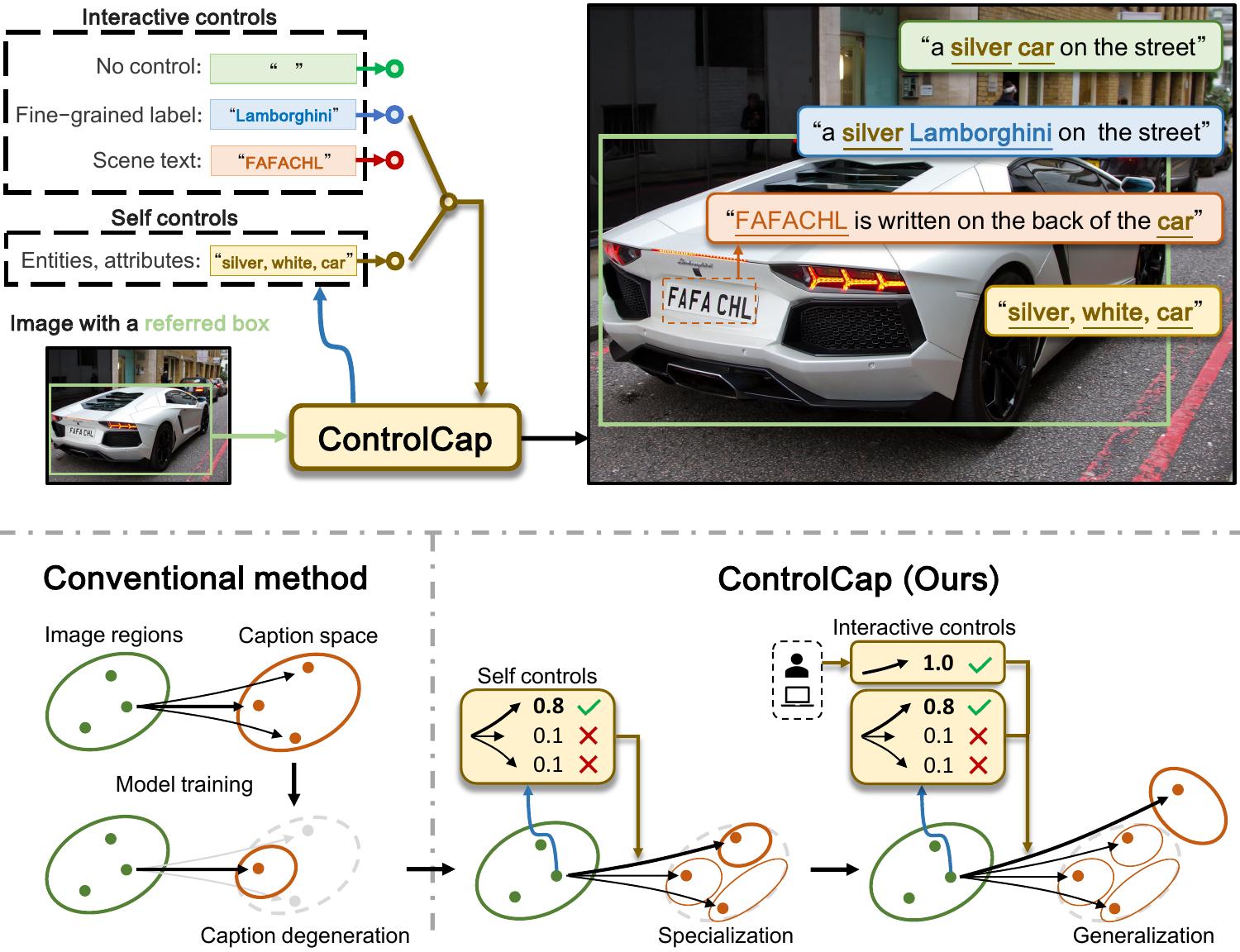}\\[-3mm]
	\caption{An illustration of \Ours (upper) and a comparison of \Ours with conventional method (lower). \Ours introduces interactive controls or self controls (such as fine-grained labels or scene text) to generate specialized captions. To generate less frequent captions, \Ours requires interactive controls such as ``Lamborghini” or ``FAFACHL”. For common captions, \Ours can generate self controls such as ``silver, white, car''. In the lower figure, the conventional method is challenged by the captioning degradation issue, $i.e.$, predicting the most frequent captions while missing the less frequent ones. In contrast, \Ours is constrained to generate captions within a few sub-spaces containing the control words so that the opportunity of hitting less frequent captions can be significant.}
    \label{fig:1}
\end{figure*}

\section{Introduction}
\label{sec:intro}

%问题分析
Region-level captioning~\cite{Johnson2016DenseCap, yu2017joint, Peng2023Kosmos2, glamm, osprey} which requires precisely describing objects within an image and completely understanding the object relations, at the same time, remains a challenging task.
%
%However, 
The key point lies that the captioning task itself is inherently ambiguous, $i.e.$, human annotators could provide totally different descriptions for an image region due to their individual intentions, while the captioning model requires to generate a consistent caption for that region. 
This ambiguity inevitably causes the caption degeneration issue~\cite{yu2023capsfusion}, $i.e.$, models predicting the most frequent captions in the training set while neglecting the less frequent ones.
The nature behind this phenomenon is that the model predictions occupy a caption space smaller than that formed by captions in the training set, Fig.~\ref{fig:1}(lower). 

%本研究解决的问题
In this study, we attempt to conquer the caption degeneration issue by breaking through the following two bottlenecks, Fig.~\ref{fig:1}(lower):
1) \textit{Specialization.}  The multimodal model is constrained to generate captions within a few sub-spaces containing the control words, so that the opportunity of hitting less frequent captions can be significant.
2) \textit{Generalization.} To maintain the diversity of captions, the trained model should be extended to accept interactive controls specified by users or perception models so that it can produce ``expected'' outputs. For example, the model responds to controls of the fine-grained label (``\texttt{Lamborghini}'') or scene text (``\texttt{FAFACHL}''), Fig.~\ref{fig:1} (upper). 
%

\begin{figure*}[t]
	\includegraphics[width=1\linewidth]{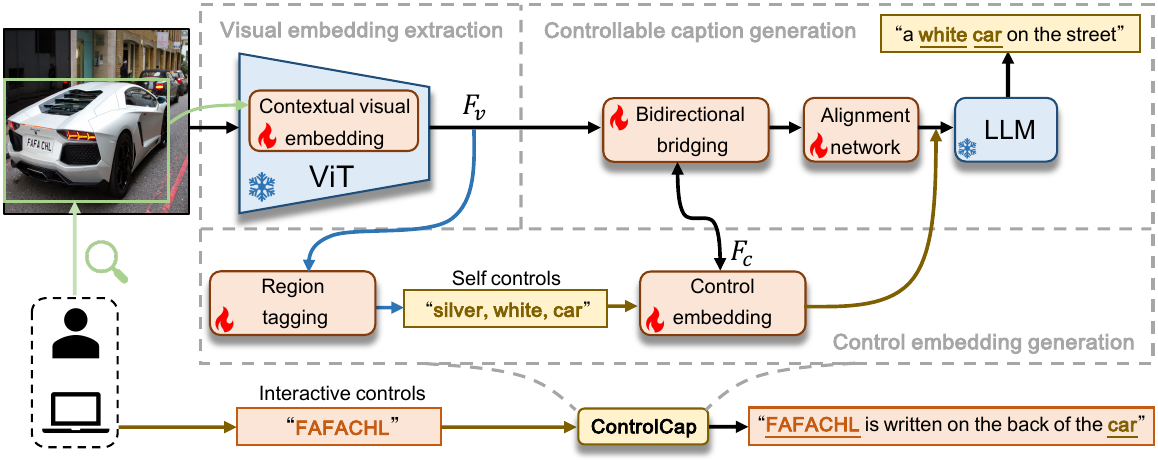}
    \caption{
    Diagram of \Ours. It comprises visual embedding extraction, control embedding generation, and controllable caption generation. visual embedding extraction consists of a frozen ViT and a contextual visual embedding module, which are introduced to enforce LMM's capacity for region-aware understanding. Control embedding generation consists of a region tagging module and a control embedding module, which are introduced to encode self controls/interactive controls. In controllable caption generation, a bidirectional bridging module maximizes the information exchange between the visual embedding $F_v$ and control embedding $F_c$. The two embeddings are then inputted into a LLM to generate specialized captions. 
    }
% \vspace{-0.1cm}
\label{fig:flowchart}
\end{figure*}

%具体方案
We propose controllable region-level captioning (ControlCap), a specific and generalizable approach to predict region-level expressions, through drawing inspirations from large multimodal models (LMMs)~\cite{Peng2023Kosmos2, Li2023BLIP2, Liu2023LLaVa} and controllable text generation methods~\cite{Zhang2022ContralTextGen, tag2text, Hu_2023_ICCV}. \Ours comprises three main components: visual embedding extraction, control embedding generation, and controllable caption generation, Fig.~\ref{fig:flowchart}.
For visual embedding extraction, a contextual visual embedding module employs two parallel and efficient branches, which balance the detail and contextual information of an image region without increasing the computation overhead. One branch captures detailed and context-free features. The other captures contextual but less detailed features, which are then merged as the visual embedding ($F_v$ in Fig.~\ref{fig:flowchart}) for caption generation.
For control embedding generation, the extracted visual embedding is fed to a region tagging module to predict corresponding control words ($i.e.$, classification categories). The control words are then encoded into the control embedding ($F_c$ in Fig.~\ref{fig:flowchart}).
The produced visual and control embedding are integrated and fed to a large language model (LLM) for controllable caption generation, Fig.~\ref{fig:flowchart}. To alleviate the variation of control words, we further introduce a bidirectional bridging module, which maximizes the information exchange between the visual embedding and the control embedding.

The contributions of this study are summarized as follows:
\begin{itemize}
    \item We propose a controllable region-level captioning (ControlCap) approach, defining a systematic way to address the caption degeneration issue by introducing control words (interactive controls and/or self controls).

    \item We design a modularized diagram, which can fully exchange information between the visual embedding and the control embedding through a bidirectional embedding bridging module, improving the accuracy of region-level captioning. 
    
    \item On Visual Genome and RefCOCOg datasets, \Ours respectively improves the CIDEr score by 21.6 and 2.2, outperforming the state-of-the-arts by significant margins.
    
\end{itemize}

\section{Related Works}
\label{sec:relate_work}

%-------------------------------------------------------------------------
\textbf{Large Multimodal Model.}
To harness the zero-shot and reasoning capabilities of large language models (LLMs)~\cite{Zhang2022OPT, Chung2022Flan5, Touvron2023LLaMA, ChatGPT2022, Brown2020GPT3}, there is a trend towards fusing vision-and-language models with LLMs to produce large multimodal models (LMMs).
Benefit from powerful foundation models~\cite{DBLP:conf/iclr/DosovitskiyB0WZ21, Devlin2018BERT, Zhang2022OPT, Chung2022Flan5} and huge amount of vision language data corpus, LMMs have achieved unprecedented performance on few-shot learning~\cite{Alayrac2023Flamingo}, visual question answering (VQA)~\cite{Li2022BLIP, Li2023BLIP2, Dai2023InstructBLIP, Liu2023LLaVa} , image captioning~\cite{Li2022BLIP, Li2023BLIP2, Dai2023InstructBLIP, Liu2023LLaVa}. 

\noindent\textbf{Region-level captioning.} This technique aims to generate detailed text descriptions for given regions. Recently, leveraging the unparalleled visual-language comprehension capabilities of large multimodal models (LMMs), the generation of region-level captions based on LMMs has become a widespread practice. Shikra~\cite{Chen2023Shikra}, GPT4RoI~\cite{zhang2023gpt4roi}, Kosmos-2~\cite{Peng2023Kosmos2}, ASM~\cite{wang2023allseeing}, MiniGPT-v2~\cite{chen2023minigptv2}, RegionGPT~\cite{guo2024regiongpt}, Alpha-CLIP~\cite{sun2023alpha}, GLaMM~\cite{glamm}, and Osprey~\cite{osprey} have enabled LMMs to achieve region-based image understanding.
They have achieved SOTA performance on region-level captioning~\cite{Chen2023Shikra, Peng2023Kosmos2, glamm, osprey, sun2023alpha}. However, suffering from the caption degeneration issue, millions of training data are required to maintain their caption space during inference. To solve this, we propose to use a discriminative module to generate control words within the caption space to divide it into multiple sub-spaces, with which the less frequent caption subspace can be highlighted by the corresponding control words, thus alleviating the degeneration issue. 

Dense captioning is a task closely associated with region-level captioning. Its objective is to identify and produce detailed descriptions for densely populated object regions within an image~\cite{Johnson2016DenseCap,Li2019Learning,Shao2022Region,Wu2022GRIT,Long2023CapDet}. As a pioneered method, FCLN~\cite{Johnson2016DenseCap} used a localization network to locate regions and a recurrent network to generate captions. JIVC~\cite{Yang2017Dense} argues that visual concepts are associated with each other. Based on the Faster R-CNN~\cite{Johnson2016DenseCap} detector, JIVC fuses image context feature with RoI (Regions of Interest) features and inferences the location and caption of objects with two LSTM~\cite{hochreiter1997long}. COCG~\cite{Li2019Learning} took a further step to fuse context features of objects in the image with RoI features. CAG-Net~\cite{Yin2019Context} introduced the features of neighboring regions and global images into the target region to generate captions for the target. 

With the advancement of transformer models, there has been a significant improvement in scene captioning~\cite{Shao2022Region,Wu2022GRIT,Long2023CapDet}.  TDC~\cite{Shao2022Region} introduced a transformer-based end-to-end architecture that leverages object relationships within images for caption decoding. GRiT~\cite{Wu2022GRIT} treats object categories as brief captions, advocating for a unified training approach for object detection and captioning models. CapDet~\cite{Long2023CapDet} combined dense captioning with open-world detection in a pretraining setup, first merging object categories with extended text definitions for alignment with RoI embeddings. Despite the progress, current methods cannot generate cross-domain captions, which limits their applicability in real-world scenarios, such as scenes that contain rich scene text. To overcome the weakness, we enable \Ours the capability of generating cross-domain captions by using interactive controls from other domains ($i.e.$, recognized scene text).

\noindent\textbf{Controllable Text Generation.}
Natural language generation (NLG) primarily aims to exert control over the text generation process by incorporating additional conditions. There are various tasks involving CTG, including attribute-based generation~\cite{DBLP:conf/iclr/DathathriMLHFMY20, DBLP:conf/nips/LiTGLH22, Carlsson2022CTG}, dialogue generation~\cite{DBLP:conf/acl/00020ZZ020}, storytelling~\cite{DBLP:conf/acl/LewisDF18}, debiasing~\cite{DBLP:conf/aaai/LiuJWXWV21}, and format control~\cite{DBLP:conf/acl/LiZLS20}.
A task closely related to ours is lexicon-controlled text generation, a form of attribute-based generation aimed at producing text focused on a specified keyword, ensuring its presence in the output~\cite{Carlsson2022CTG,DBLP:conf/nips/LiTGLH22}.
Existing studies implemented achieve lexicon control through techniques like fine-tuning~\cite{Carlsson2022CTG}, post-processing~\cite{DBLP:conf/iclr/DathathriMLHFMY20}, and diffusion~\cite{DBLP:conf/nips/LiTGLH22}.
For image captioning and tagging, LaNAR~\cite{ding2023image} tried providing image captions with specified levels of detail by managing the length of generated captions. PromptCap~\cite{Hu_2023_ICCV} and Tag2Text~\cite{tag2text} leveraged natural language prompts to direct the description of visual entities in the generated captions.

Existing studies have the capability to produce fluent text that meets certain conditions or controls the generated captions at image-level. 
Nevertheless, the capability to produce specialized captions for designated regions remains unsolved.

\section{The Proposed Approach}
\label{sec:method}

\subsection{Overview}
\Ours leverages a pre-trained large multimodal model composed of a frozen vision transformer~\cite{Radford2021CLIP} (ViT), an alignment network, and a frozen large language model (LLM), ig.~\ref{fig:flowchart}. 
To achieve controllable region-level captioning, \Ours proposes visual embedding extraction, control embedding generation, and controllable caption generation, Fig.~\ref{fig:flowchart}.
For the visual embedding extraction, a contextual visual embedding module collaborates with the ViT to extract a visual embedding $F_v$ from a given image region (Sec.~\ref{sec:vee}). 
Then in control embedding generation, the extracted visual embedding is fed to a region tagging module to predict control words $c$, which are then fed into an embedding module to generate a control embedding $F_c$ (Sec.~\ref{sec:ceg}).
Finally, the produced visual embedding and control embedding exchange information via a bidirectional bridging module to reduce the misalignment issue caused by various controls. The visual embedding is projected into the language feature space by the alignment network, which is then fed to the LLM together with the control embedding for controllable caption generation (Sec.~\ref{sec:ccg}).

Let $x$ denote a training image. $b$ denotes a referred box. $y$ denotes the ground-truth caption corresponding to $b$. The training loss of \Ours is defined as
\begin{equation}
    {\cal L}_{\text{ControlCap}}(x, b, y) 
    = {\cal L}_{\text{tag}} (x, b, {\cal C}_t(y)) 
    + {\cal L}_{\text{cap}} (x, b, {\cal C}_l(y), y)
\end{equation}
${\cal L}_{\text{tag}}$ indicates the tagging loss~\cite{ridnik2021asymmetric} (added atop the region tagging module) and ${\cal L}_{\text{cap}}$ the captioning loss~\cite{Li2023BLIP2} (added atop the LLM). 
${\cal C}_t(y)$ and ${\cal C}_l(y)$ are control words generated through extracting informative words from $y$ (detailed in Sec.~\ref{sec:ceg}), while they respectively denote the ground-truths for the tagging loss and the control words for the captioning model.
During inference (Sec.~\ref{sec:inference}), by giving an image $x$ and a referred box $b$, \Ours generates specialized captions under interactive controls $c$ (bottom of Fig.~\ref{fig:flowchart}), which can be given by users or perception models. 

\subsection{Visual Embedding Extraction}
\label{sec:vee}
% \label{sec:meg}

\begin{figure}[t]
	\includegraphics[width=0.98\linewidth]{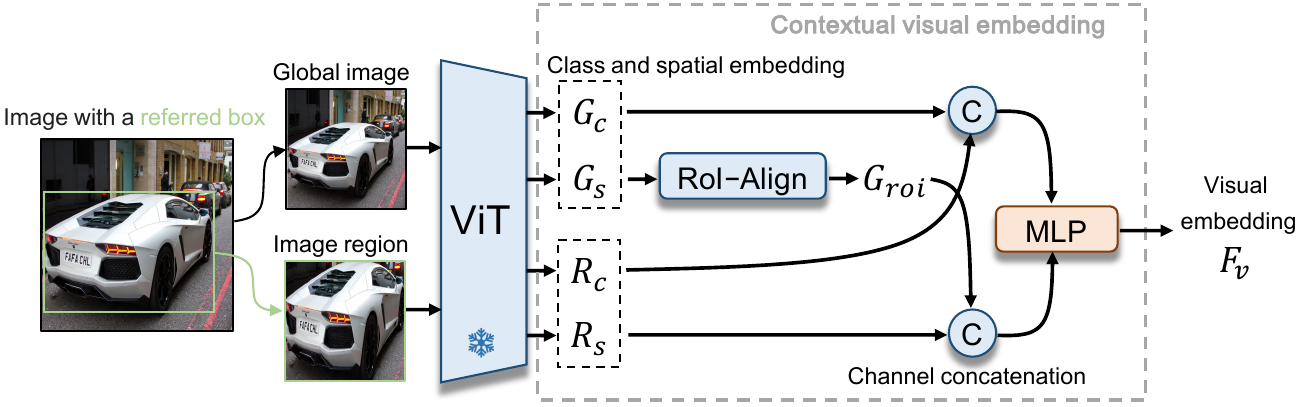}
	\caption{Diagram for visual embedding extraction.}
\label{fig:cve_arc}
\end{figure}

% motivation
For region-level visual tasks ($e.g.$, object detection~\cite{he2017mask}, dense captioning~\cite{Johnson2016DenseCap}), the model requires not only the ability to discern details within an image region but also to perceive the overall image context. However, constrained by the high computational cost of large multimodal models, existing methods~\cite{Wu2022GRIT, Peng2023Kosmos2, zhang2023gpt4roi, glamm} are limited to using low-resolution image inputs, which can degrade image details, particularly for small regions. One simple approach to enhance region details involves extracting embeddings from upscaled and cropped image regions. Nonetheless, this approach cannot perceive the overall image context.
To solve the conflict between detail-rich visual embedding and the computational overhead brought by the details, we design a contextual visual embedding module to extract and merge two parallel and specialized embeddings, Fig.~\ref{fig:cve_arc}.

Initially, the image $x$ is scaled down to a lower resolution (global image) and inputted into the ViT, where it is encoded into a class embedding $G_c$ and a spatial embedding $G_s$. Subsequently, a RoI-align module~\cite{he2017mask} extracts the RoI embedding $G_{roi}$ that is context-aware and facilitates faster computation.
We then crop an image region according to the location of a referred box $b$, which is resized to the same size as the global image and fed to the ViT to extract a detail-rich class embedding $R_{c}$ and a spatial embedding $R_s$.
To couple the context information, we concatenate $G_c$ and $R_c$, $G_s$ and $R_s$ across channel dimensions respectively, followed by passing through a learnable multi-layer perceptron (MLP) module so that we extract the visual embedding $F_v$ for the referred box $b$ by merging the output embeddings of the MLP.

\subsection{Control Embedding Generation}
\label{sec:ceg}
% motivation
We then utilize the extracted visual embedding $F_v$ to further generate the control embedding $F_c$. To ensure that $F_c$ can be employed to address the caption degeneration issue while also ensuring generalization to new domains of captions, the control words that are used to generate $F_c$ need to satisfy the following challenging conditions: 
(1) these control words should be able to reduce the ambiguity in the vision-caption mapping relationship caused by the diversity of captions; 
(2) these control words need to be adaptively obtained based on the visual content within the region, by the model itself or be specified by humans or expert models; 
(3) these control words should cover the caption space as much as possible during training to improve the model’s generalization ability.

\begin{figure}[t]
	\includegraphics[width=1.0\linewidth]{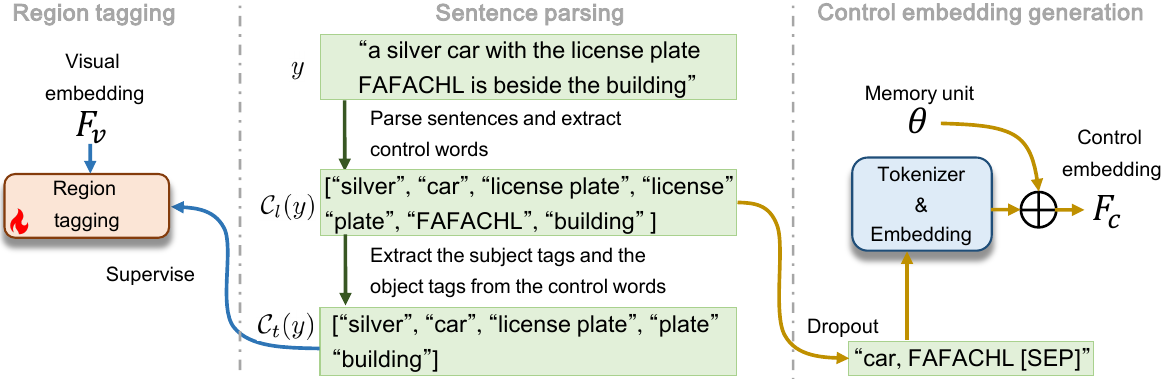}
	\caption{Diagram of control embedding generation during training.}
\label{fig:cee_arc}
\end{figure}

To address the first condition, we innovatively introduce a discriminative model (region tagging module) to predict these control words. Unlike caption models, the predictions of discriminative models are typically unambiguous (as the annotations for discriminative models tend to be unambiguous). Therefore, controlling the caption generation process using the predictions of the discriminative model can reduce ambiguity issues.
For the second condition, we use the visual features $F_v$ as the input to this discriminative model, to predict control words relevant to the region. As presented in the last subsection, $F_v$ simultaneously captures detailed information within the region and global context information around, ensuring that the discriminative model has the potential to output a more comprehensive range of control words including those from humans or expert models.
To address the third condition, we parse the ground-truth captions into ground-truth control words, which are utilized to supervise the discriminative model, Fig.~\ref{fig:cee_arc}. Since control words are guaranteed to appear in ground-truth captions, controlling based on these words ensures their presence in the output captions as much as possible.
Through this approach, each control word partitions a subspace from the caption space. Given special control words, the caption degeneration issue is alleviated.

\noindent\textbf{Discriminative model:} We adopt the recently popularly used tagging method as the discriminative model. Inspired by the queried-based image tagging methods~\cite{query2label, tag2text, ram}, we apply a lightweight recognition decoder~\cite{query2label} to generate visual-related tags within a region. Following~\cite{ram}, we utilize a class set of 4585 classes, ranging from entities, attributes, actions, and scenes, which supports the caption space.

During training, we get the region tags ${\cal C}_t(y)$ by parsing the ground-truth caption $y$ into control words and filtering the words that are not in the class set, Fig.~\ref{fig:cee_arc} (middle). However, the caption might include some less related concepts outside the region, which makes the training of the region tagging module unstable. To solve that, we split the region tags into two disjoint subsets, including the subject tag set and the object tag set. The subject tag set (${\cal C}^s_t(y)$) contains the subject along with its adjectives and adverbs of the caption, which usually appear in the region. The object tag set contains other tags that are related to the region, $i.e.$, ${\cal C}_t(y)-{\cal C}^s_t(y)$. These two sets are used to jointly supervise the region tagging module of $4585\times 2$ classes, Fig.~\ref{fig:cee_arc} (left). Due to the presence of missing labels in regions, asymmetric loss~\cite{ridnik2021asymmetric} is used for optimization.

\noindent\textbf{Control embedding:} 
The control words are encoded to control embedding so that the LLM can take it as input and generate specialized captions about an image region.
During training, control words are randomly dropped in accordance with a Bernoulli distribution. The remained control words are shuffled and combined with a $\texttt{[SEP]}$ token to form a control sentence, Fig.~\ref{fig:cee_arc} (right). We utilize the tokenizer and word embedding layer of the LLM to encode the sentence into the control embedding.
We further develop a memory unit that uses a 1D learnable parameter $\theta \in \mathbb{R}^{D}$ to guarantee generalized controllable ability with the empty string ($i.e.$, all control words are dropped). $D$ is the dimension of control embeddings. The control embeddings are then updated by adding each of them with $\theta$.

\subsection{Controllable Caption Generation}
\label{sec:ccg}

\begin{figure}[t]
	\includegraphics[width=0.98\linewidth]{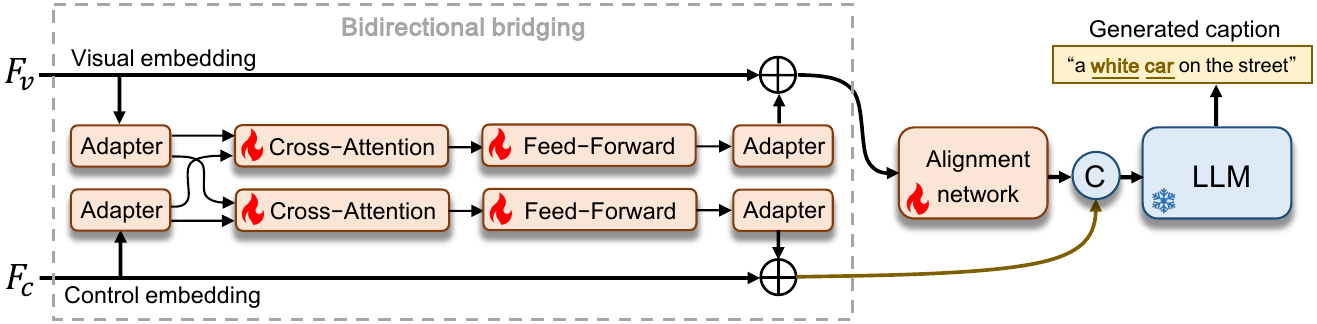}
	\caption{Diagram of the bidirectional bridging module, which maximizes the information exchange between the visual embedding and control embedding modules.}
\label{fig:beb_arc}
\end{figure}

After control embedding generation, the produced visual embedding and control embedding are fed to the LLM for controllable caption generation.
% motivation
However, for each visual embedding, there might be multiple control embeddings encoded by different control words. It is hard to align all these control embeddings to a single visual embedding, which we refer to as the variation issue of control words.
To alleviate the variation issue, we design a bidirectional bridging (BiB) module to maximize the information exchange between the visual embedding and control embedding for better alignment between them, Fig.~\ref{fig:beb_arc}.

BiB module is composed of three types of layers, $i.e.$, adapter layers,  cross-attention layers~\cite{vaswani2017attention} and feed-forward layers~\cite{vaswani2017attention}.
Adapter layers are single linear layers that aim to map the visual embedding $F_v$ or the control embedding $F_c$ to a low-dimensional latent space, Fig.~\ref{fig:beb_arc} (left), or map them back to the original feature space, Fig.~\ref{fig:beb_arc} (right).
Visual and control embeddings are first mapped to the same latent space by two adapter layers.
Features from the control embedding are then transmitted to the visual embedding by a cross-attention layer and a feed-forward layer, which uses control embedding as $\mathrm{Key, Value}$ and visual embedding as $\mathrm{Query}$, Fig.~\ref{fig:beb_arc} (upper).
Meanwhile, features from the visual embedding are transmitted to the control embedding by a cross-attention layer and a feed-forward layer, which use control embedding as $\mathrm{Query}$ and visual embedding as $\mathrm{Key, Value}$, Fig.~\ref{fig:beb_arc} (bottom).
Finally, the feature-enhanced visual and control embeddings are mapped back to their original feature space and fused with the original ones through residual connections~\cite{he2016deep}.

\subsection{Controllable Inference}
\label{sec:inference}
With a trained \Ours model, we can perform controllable inference in specialized scenarios, Fig.~\ref{fig:flowchart} (bottom). 
Before inference, users or models can specify the regions and the control words ($e.g.$, SAM~\cite{sam}, text spotting models, object detection models). The interactive controls and predicted self controls are uniformly encoded as the control embedding. \Ours then produces captions for specialized scenarios.

\section{Experiment}
\label{sec:experiment}

\noindent\textbf{Implementation Details.}  \Ours is implemented upon the LAVIS~\cite{li2022lavis} framework, where ViT, LLM and alignment network are respectively implemented using 
EVA~\cite{fang2023eva}, Flan-T5$_{\text{XL}}$~\cite{Chung2022Flan5} and Q-former~\cite{Li2023BLIP2}, Fig.~\ref{fig:flowchart}.
The models are trained using 8 NVIDIA A800 GPUs, with the Adam optimizer where the batch size is set to 768.
Without otherwise specified, all models are trained by 5 epochs and the initial learning rate is set to $1 \times 10^{-4}$ with a cosine learning rate decay.
During inference, the beam size of the LLM is set to 3 and a single caption is generated for each referred region.

\noindent\textbf{Datasets.} For dense captioning, \Ours is trained using VG or VG-COCO~\cite{Shao2022Region}. For referring expression generation, \Ours is trained using Visual Genome (VG)~\cite{krishna2017visual} and RefCOCOg~\cite{yu2016modeling}. 
VG dataset is a finely labeled dataset with dense annotations of objects, attributes, and relationships. VG-COCO~\cite{Shao2022Region} is the intersection of VG V1.2 and MS COCO~\cite{lin2014microsoft}. RefCOCOg contains relatively long descriptions that describe the specific regions from various perspectives.

\noindent\textbf{Evaluation Metrics.} We follow the setting of ~\cite{Johnson2016DenseCap, Peng2023Kosmos2} to evaluate the dense captioning performance of \Ours on VG, VG-COCO and the referring expression generation performance of \Ours on VG, RefCOCOg. For dense captioning, mean Average Precision (mAP)~\cite{Johnson2016DenseCap} is adopted as the evaluation metric. the mAP is calculated across a range of thresholds for both localization and language accuracy, $i.e.$, the intersection over union (IOU) thresholds (0.3, 0.4, 0.5, 0.6, 0.7) are used for localization and the METEOR score’ thresholds (0, 0.05, 0.1, 0.15, 0.2, 0.25) is adopted for evaluating the language generation. Since \Ours lacks the capability to perform object detection, we utilize a GRiT~\cite{Wu2022GRIT} model trained on VG to acquire object locations. 

To evaluate the region-level captioning performance without being affected by the localization performance, we also evaluate the model when ground-truth bounding boxes are given during inference.
For referring expression generation, we adopt the METEOR score and CIDEr score to evaluate the caption quality of \Ours. 
Different from the previous methods, \Ours can generate specialized captions given interactive controls. To evaluate such ability, the first noun in the ground-truth caption is used to simulate the interactive control during inference (``interactive control'' in Tabs.~\ref{tab:ablation_module} and ~\ref{tab:ablation_beb}). For example, for the caption ``\texttt{a black car is parked beside the street}'', the word ``\texttt{car}'' is provided to \Ours as the interactive control.

\begin{table}[t]
    \centering
    \tabcolsep=0.15cm
    \begin{tabular}{rc|c|c|c}
\toprule
\multirow{2}{*}{Methods} & \multirow{2}{*}{GT localization} & \multicolumn{3}{c}{mAP(\%)}\tabularnewline
\cline{3-5} 
 &  & VG V1.0 & VG V1.2 & VG-COCO\tabularnewline
\midrule
FCLN$_{\text{CVPR'16}}$~\cite{Johnson2016DenseCap} & \ding{55} & 5.4 & 5.2 & -\tabularnewline
JIVC$_{\text{CVPR'17}}$~\cite{Yang2017Dense} & \ding{55} & 9.3 & 10.0 & -\tabularnewline
ImgG$_{\text{AAAI'19}}$~\cite{Li2019Learning} & \ding{55} & 9.3 & 9.7 & -\tabularnewline
COCD$_{\text{AAAI'19}}$~\cite{Li2019Learning} & \ding{55} & 9.4 & 9.8 & 7.9\tabularnewline
COCG$_{\text{AAAI'19}}$~\cite{Li2019Learning} & \ding{55} & 9.8 & 10.4 & 8.9\tabularnewline
CAG-Net$_{\text{CVPR'19}}$~\cite{Yin2019Context} & \ding{55} & 10.5 & - & -\tabularnewline
TDC$_{\text{TNNLS'22}}$~\cite{Shao2022Region} & \ding{55} & 11.5 & 11.9 & 11.9\tabularnewline
GRiT$_{\text{ARXIV'22}}$~\cite{Wu2022GRIT} & \ding{55} & 15.5 & 16.4 & -\tabularnewline
CapDet$_{\text{CVPR'23}}$~\cite{Long2023CapDet} & \ding{55} & - & 15.4 & 14.0\tabularnewline
DCMSTRD$_{\text{TMM'24}}$~\cite{10444947} & \ding{55} & 13.6 & 13.4 & 16.1\tabularnewline
% \rowcolor{gray0} \Ours (Ours) & \ding{55} & \textbf{17.6} & \textbf{18.0} & \textbf{17.7}\tabularnewline
% \midrule v1 version
\rowcolor{gray0} \Ours (Ours) & \ding{55} & \textbf{18.2} & \textbf{18.5} & \textbf{18.4}\tabularnewline
\midrule
FCLN$_{\text{CVPR'16}}$~\cite{Johnson2016DenseCap} & \ding{51} & 27.0 & - & -\tabularnewline
JIVC$_{\text{CVPR'17}}$~\cite{Yang2017Dense} & \ding{51} & 33.6 & - & -\tabularnewline
CAG-Net$_{\text{CVPR'19}}$~\cite{Yin2019Context} & \ding{51} & 36.3 & - & -\tabularnewline
GRiT$_{\text{ARXIV'22}}$~\cite{Wu2022GRIT} & \ding{51} & 40.0 & 40.3 & -\tabularnewline
BLIP2$_{\text{ICML'23}}$~\cite{Li2023BLIP2} & \ding{51} & 37.7 & 37.9 & 36.9\tabularnewline
\rowcolor{gray0} \Ours (Ours) & \ding{51} & \textbf{42.4} & \textbf{42.8} & \textbf{43.2}\tabularnewline
\bottomrule
\end{tabular}
    \caption{Comparison of dense captioning performance of the proposed approach with the state-of-the-art methods on the VG and VG-COCO datasets.}
    \label{tab:performance_vg_dense_caption}
\end{table}

\begin{table}[t]
    \centering
    \tabcolsep=0.15cm
    \begin{tabular}{rc|c|c|c|c}
\toprule
\multirow{2}{*}{Method} & \multirow{2}{*}{Model size} & \multicolumn{2}{c|}{RefCOCOg} & \multicolumn{2}{c}{VG}\tabularnewline
\cline{3-6} 
 &  & METEOR & CIDEr & METEOR & CIDEr\tabularnewline
\midrule
SLR+Rerank$_{\text{CVPR'17}}$~\cite{yu2017joint} & <1B & 15.9 & 66.2 & -  & -\tabularnewline
GRiT$_{\text{ARXIV'22}}$~\cite{Wu2022GRIT} & <1B & 15.2 & 71.6 & 17.1 & 142.0\tabularnewline
Kosmos-2$_{\text{ICLR'24}}$~\cite{Peng2023Kosmos2} & 1.6B & 14.1 & 62.3 & -  & -\tabularnewline
GPT4RoI$_{\text{ARXIV'23}}$~\cite{zhang2023gpt4roi} & 7B & - & - & 17.4 & 145.2\tabularnewline
RegionGPT$_{\text{CVPR'24}}$~\cite{glamm} & 7B & 16.9 & 109.9 & 17.0 & 145.6\tabularnewline
GLaMM$_{\text{CVPR'24}}$~\cite{glamm} & 7B & 16.2 & 105.0 & 18.6 & 157.8\tabularnewline
Alpha-CLIP+LLaVA$_{\text{CVPR'24}}$~\cite{sun2023alpha} & 7B & 16.7 & 109.2 & 18.9 & 160.3\tabularnewline
Osprey$_{\text{CVPR'24}}$~\cite{osprey} & 7B & 16.6 & 108.3 & - & -\tabularnewline
\midrule
\rowcolor{gray0} \Ours (Ours) & 4.2B & \textbf{17.0} & \textbf{111.4} & \textbf{20.4} & \textbf{181.9}\tabularnewline
\rowcolor{gray0} \textcolor{gray}{\Ours$\dagger$ (Ours)} & \textcolor{gray}{4.2B} & \textcolor{gray}{\textbf{21.3}} & \textcolor{gray}{\textbf{168.7}} & \textcolor{gray}{\textbf{28.8}} & \textcolor{gray}{\textbf{302.3}}\tabularnewline
% \rowcolor{gray0} \Ours$_{\text{Vicuna-7B}}$ (Ours) & 8B &  &  &  & \tabularnewline
\bottomrule
\end{tabular}

    \caption{Referring expression generation performance of the proposed approach and the state-of-the-art methods on the RefCOCOg and VG datasets. $\dagger$ denotes that the first noun in the ground-truth caption is used to simulate the interactive control.}
    \label{tab:performance_reg}
\end{table}

\subsection{Performance}

\noindent\textbf{Dense Captioning.} 
In Tabs.~\ref{tab:performance_vg_dense_caption}, the dense captioning performance of \Ours is compared with the state-of-the-art (SOTA) methods.
\Ours respectively achieves 18.2\%, 18.5\% and 18.4\% mAPs on VG V1.0, VG V1.2, and VG-COCO, outperforming the SOTA methods by significant margins.
When ground-truth bounding boxes are given, \Ours respectively achieves 42.4\%, 42.8\% and 43.2\% mAPs on VG V1.0, VG V1.2, and VG-COCO, outperforming BLIP2~\cite{Li2023BLIP2} by 6.3\% on VG-COCO.
% while introducing only 0.18\% parameters (Fig.~\ref{fig:model_param}).
% %
% Last but not least, when linguistic guidance is given, \Ours respectively achieves 65.81\% mAP and 66.12\% mAP on V1.0 and V1.2, which implies that linguistic guidance can greatly boost the performance of dense captioning.

\noindent\textbf{Referring Expression Generation.} In Tabs.~\ref{tab:performance_reg}, the referring expression generation performance of \Ours is compared with the SOTA methods. \Ours respectively achieves 17.0 and 20.4 METEOR scores, 111.4 and 181.9 CIDEr scores on RefCOCOg and VG, outperforming the SOTA methods with a much smaller model size (4.2B $vs.$ 7B). 
we simulate the performance of \Ours under interactive controls by using the first noun in the ground-truth caption as control words. \Ours achieves a 28.8 METEOR score and 302.3 CIDEr score on VG under this condition. 

\begin{table}[t]
    \centering
    \small
    \begin{tabular}{c|ccc}
\toprule
 & ImageNet-1K & Object365 & ICDAR2015\tabularnewline
\midrule
Control accuracy  & 87.3\% & 95.5\% & 82.4\%\tabularnewline
GPT-4v preference & (80 / 8 / 12) & (54 / 5 / 41) & (73 / 6 / 21)\tabularnewline
\bottomrule 
\end{tabular}
    \caption{Evaluation of the controllable ability of \Ours under specialized scenes. Control accuracy is defined as the proportion of successfully controlled captions to all captions. A successfully controlled caption is supposed to contain the word used to control. Human study under various scenarios. GPT-4v is employed to mimic human preferences for captions generated by ControlCap. $(N_1 / N_2 / N_3)$ indicates the frequency with which GPT-4v assesses that (controlled caption is better / uncontrolled caption is better / both captions are of equal quality).}
    \label{tab:human_study}
\end{table}

\begin{figure*}[t]
	\includegraphics[width=0.72\linewidth]{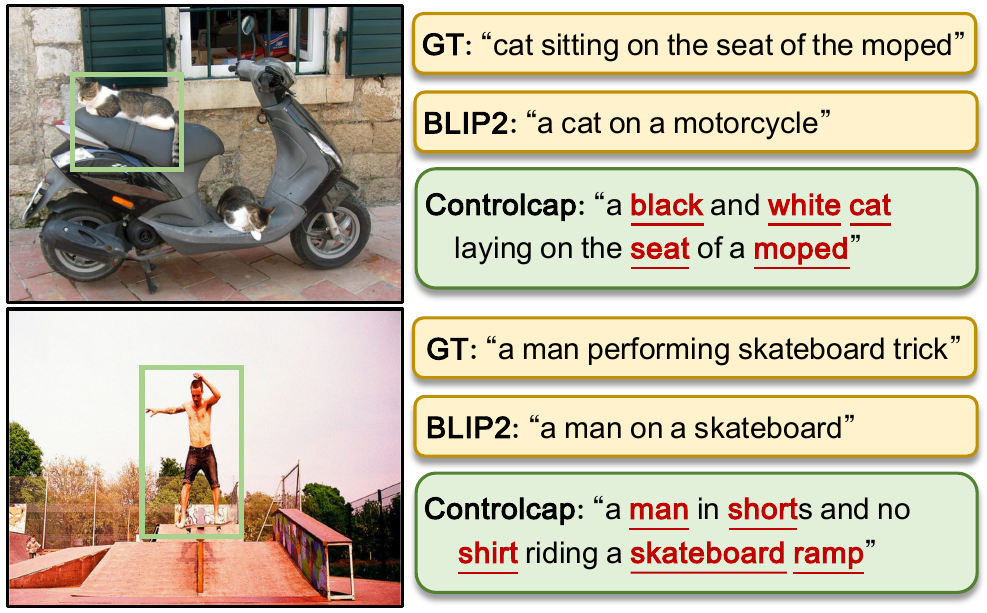}
	\caption{Qualitative comparison of the ground-truth (GT) captions on RefCOCOg, BLIP2 and \Ours. The red underlined words are the generated self controls.}
\label{fig:exp_fig1}
\end{figure*}

\begin{figure*}[th]
	\includegraphics[width=0.98\linewidth]{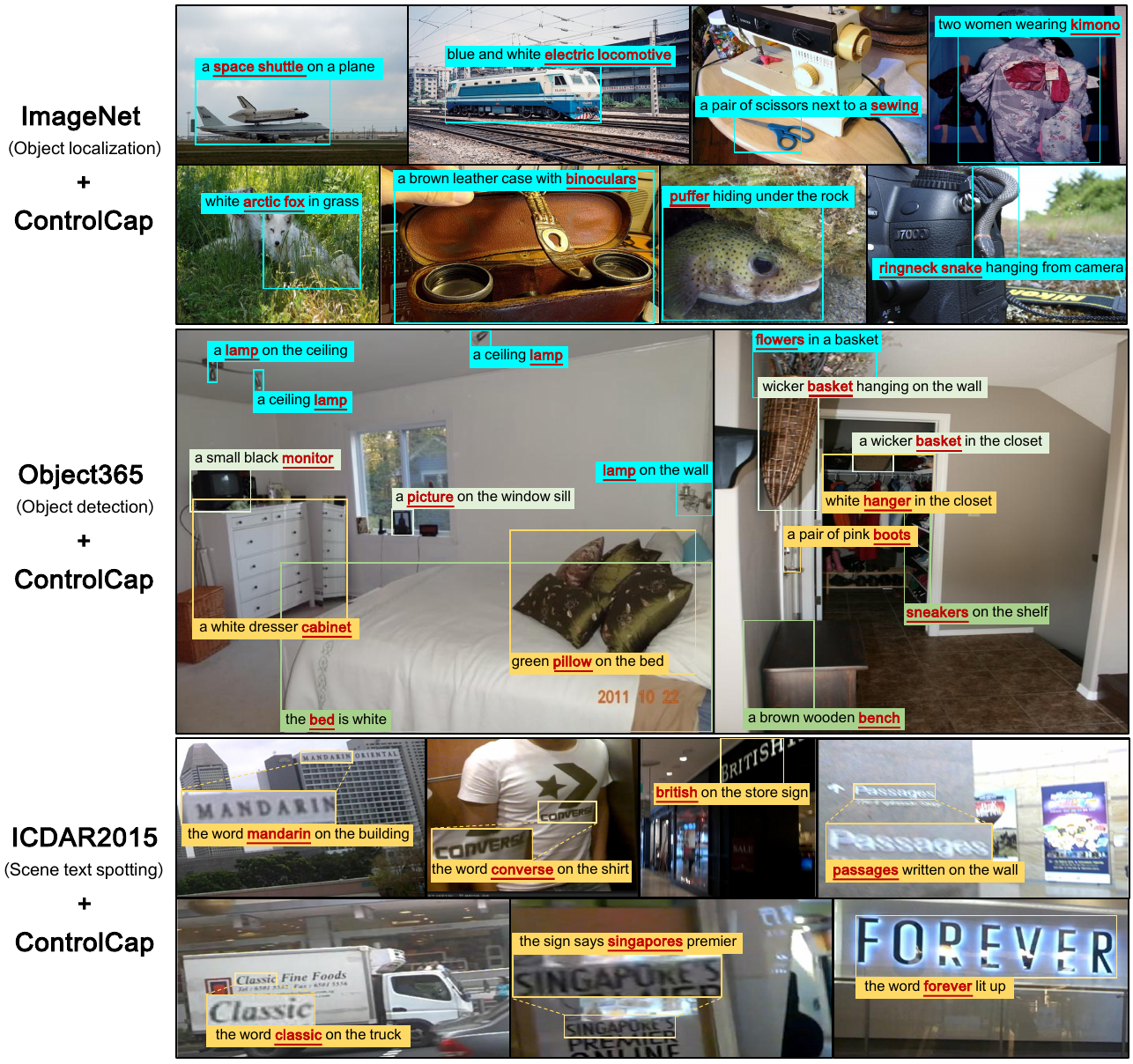}
    \caption{Qualitative analysis of the cross-domain captioning capabilities of \Ours. By combining pre-trained \Ours with datasets that either contain fine-grained category labels ($e.g.$, ImageNet used for object localization, ICDAR2015 used for text spotting) or abundant samples ($e.g.$, Object365 used for object detection), specialized captions can be generated. The red underlined words are used as the interactive controls.}
\label{fig:exp_fig2}
\end{figure*}

\noindent\textbf{Controllable Inference.} We evaluate the controllable ability using three vision tasks, including object localization on ImageNet-1K~\cite{deng2009imagenet}, object detection on Object365~\cite{shao2019objects365}, and scene text spotting on ICDAR2015~\cite{karatzas2015icdar}. In these tasks, by receiving object categories (scene text) as control words, \Ours generates specialized captions for each image region. We first evaluate control accuracy to check whether the caption contains the control words (Successful control) or not (unsuccessful control). As shown in Tab.~\ref{tab:human_study} first row, the control accuracy is consistently higher than 80\%, which indicates that \Ours is capable of generating specialized captions under different settings.

We also evaluate the effect of controls by comparing the captions with those generated without interactive control words. We utilized GPT-4v as an objective and impartial agent to judge the quality of the two kinds of captions, Tab.~\ref{tab:human_study} second row. We provide GPT-4v 100 images with white rectangular borders highlight the region for each scenario and use prompt ``\textit{Caption1: \{\texttt{cap1}\}, Caption2: \{\texttt{cap2}\}. Please compare the professionalism and accuracy of the two captions based on the white rectangular region in the pictures. Choose from the following three options: 1. Caption1 is better. 2. Caption2 is better. 3. They are equally good.}'' \{\texttt{cap1}\} and \{\texttt{cap2}\} are tested captions. It can seen that the quality of captions with interactive controls is significantly better than that without control in various scenarios. 

\noindent\textbf{Qualitative Visualizations.} 
Fig.~\ref{fig:exp_fig1} compares the captioning results of BLIP2 and \Ours.
Suffering from the caption degeneration issue, BLIP2 predicts simple and less informative captions.
By introducing self controls (The red underlined words in Fig.~\ref{fig:exp_fig1}), \Ours generates informative captions, which are even longer than the ground-truth annotations.

Fig.~\ref{fig:exp_fig2} demonstrates \Ours's generalization capability, $e.g.$, generating captions beyond the caption space during training under interactive controls, such as ImageNet with fine-grained category labels, Object365 with abundant region-category pairs, and ICDAR2015 with scene text.
The ability implies that \Ours can either be combined with various datasets to generate domain-specific region-caption datasets or be combined with specialist models ($e.g.$, classifier, detector, and text spotter) to form a specialized region-level captioning model.

\begin{table}[t]
    \centering
    \tabcolsep=0.2cm
    \caption{Ablation studies of the components in \Ours on VG V1.2. The first noun in the ground-truth caption is used to simulate the interactive controls. CVE, RegionTag, CE, BiB respectively denote the contextual visual embedding, the region tagging, the control embedding, and the bidirectional bridging in Fig.~\ref{fig:flowchart}.}
    \begin{tabular}{c|cccc|c|c}
\toprule
\multirow{2}{*}{} & \multirow{2}{*}{CVE} & \multirow{2}{*}{RegionTag} & \multirow{2}{*}{CE} & \multirow{2}{*}{BiB} & \multicolumn{2}{c}{mAP(\%)}\tabularnewline
\cline{6-7}
 &  &  &  &  & self control & interactive control\tabularnewline
\midrule
1 & \ding{55} & \ding{55} & \ding{55} & \ding{55} & 37.9 & -\tabularnewline
2 & \ding{51} & \ding{55} & \ding{55} & \ding{55} & 42.4 & -\tabularnewline
3 & \ding{51} & \ding{55} & \ding{51} & \ding{55} &  42.0& 65.1\tabularnewline
% 5 & \ding{51} & \ding{51} & \ding{51} & \ding{55} & 43.1 & \tabularnewline
4 & \ding{51} & \ding{55} & \ding{51} & \ding{51} & 42.4 & 65.8\tabularnewline 
5 & \ding{51} & \ding{51} & \ding{51} & \ding{51} &  42.8 & 69.0\tabularnewline
\bottomrule
\end{tabular}
    \label{tab:ablation_module}
\end{table}

\subsection{Ablation Studies}
\label{sec:ablation}

\noindent\textbf{Baseline.} The baseline model is BLIP2~\cite{Li2023BLIP2}.
We finetune the Q-former in BLIP2 on the region-caption pairs cropped from VG or VG-COCO. The performance of BLIP2 on VG and VG-COCO are shown in Tab.~\ref{tab:performance_vg_dense_caption}. It achieves 37.9\% mAP on VG V1.2.

\noindent\textbf{Visual Embedding Extraction.}
By adding the contextual visual embedding (CVE in Tab.~\ref{tab:ablation_module}), a performance gain of 4.5\% (42.4\% $vs.$ 37.9\%) can be achieved in mAP (Line 1-2 in Tab.~\ref{tab:ablation_module}), while dropping the detail-rich region features, the context-aware RoI features or the class embeddings all hurt the performance (Line 1-3 in Tab.~\ref{tab:ablation_cve}). The results imply that fusing the region features of detailed information and the context-aware RoI features can boost the performance of region-level captioning.

\noindent\textbf{Control Embedding Generation.} 
By adding the control embedding (CE in Tab.~\ref{tab:ablation_module}), the model gains the ability to generate captions under controls (Line 3 in Tab.~\ref{tab:ablation_module}). However, the performance of \Ours in mAP drops to 42.0\%, suffering from the variation issue of control words.
By adding the region tagging module (RegionTag in Tab.~\ref{tab:ablation_module}) to generate self controls, a performance gain of 0.4\% (42.8\% $vs.$ 42.4\%) can be achieved in mAP. Performance on VG under different tagging thresholds is shown in Tab.~\ref{tab:ablation_tag_thr}. A threshold of around 0.8 leads to the best result.

\noindent\textbf{Controllable Caption Generation.} 
The bidirectional bridging (BiB) module has two branches. On the one hand, the control embedding $F_c$ is enhanced by information from the visual embedding $F_r$ ($F_r\rightarrow F_c$ in Fig.~\ref{fig:beb_arc}). On the other hand, the visual embedding $F_r$ is enhanced by information from the control embedding $F_c$ ($F_c\rightarrow F_r$ in Fig.~\ref{fig:beb_arc}).
By adding the $F_c\rightarrow F_r$ branch, the mAP of \Ours improves both with self controls and with interactive controls (Line 1-2 in Tab.~\ref{tab:ablation_beb}). We visualize the cross-attention maps from the cross-attention layer in the $F_c\rightarrow F_r$ branch in Fig.~\ref{fig:viz_attn}.
The activated regions are highly correlated to the generated captions, demonstrating that the BiB module can guide the visual embedding to align with the current control embedding, thus alleviating the variation issue of control words.

By adding both the two branches, the performance of \Ours in mAP further improves (Line 3 in Tab.~\ref{tab:ablation_beb}). The results imply that aligning the control embedding with the visual embedding can increase the model's adaptability to different controls.

\begin{table*}[t]
  \small
  \begin{floatrow}
  \capbtabbox{
  \centering
    \includegraphics[width=0.98\linewidth]{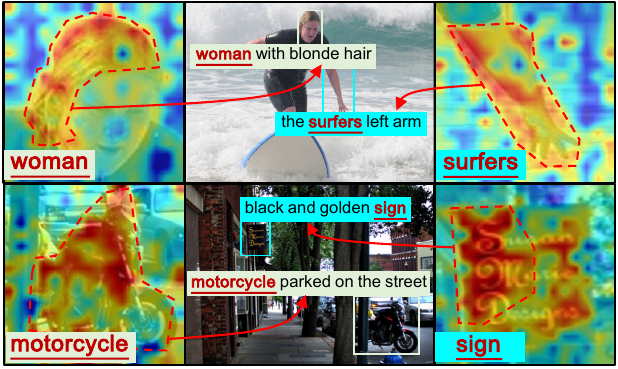}
  }
  {
  \captionof{figure}{Visualizations of attention maps from the bidirectional bridging (BiB) module. The red underlined words are used as control words.}
\label{fig:viz_attn}
  }
  \capbtabbox{
  \centering
    \includegraphics[width=0.85\linewidth]{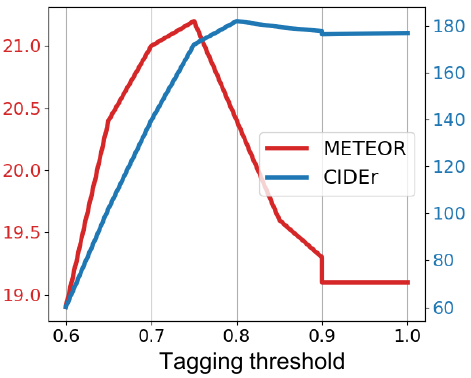}
  }
  {
  \captionof{figure}{Referring expression generation performance on VG under different tagging thresholds.}
    \label{tab:ablation_tag_thr}
  }
  \end{floatrow}
\end{table*}

\begin{table*}[t]
  \small
  \begin{floatrow}
  \capbtabbox{
  \centering
    \footnotesize
    \begin{tabular}{c|ccc|c}
\toprule
 & Region & RoI & Cls & mAP (\%)\tabularnewline
\midrule
1 & \ding{51} & \ding{55} & \ding{51} & 37.9\tabularnewline

2 & \ding{55} & \ding{51} & \ding{51} & 35.9\tabularnewline

3 & \ding{51} & \ding{51} & \ding{55} & 39.7\tabularnewline
4 & \ding{51} & \ding{51} & \ding{51} & 42.4\tabularnewline
\bottomrule
\end{tabular}
 
  }
  {
  \captionof{table}{Evaluation of contextual visual embedding module. ``Region'', ``RoI'', and ``Cls'' respectively denotes $[R_s, R_c]$, $[G_s, G_c]$, $[R_c, G_c]$ in Fig.~\ref{fig:cve_arc}.}
    \label{tab:ablation_cve} 
  }
  \capbtabbox{
  \footnotesize
    \centering
    \begin{tabular}{c|cc|c|c}
\toprule
\multirow{2}{*}{} & \multirow{2}{*}{$F_c\rightarrow F_r$} & \multirow{2}{*}{$F_r\rightarrow F_c$} & \multicolumn{2}{c}{mAP (\%)}\tabularnewline
\cline{4-5}
 &  &  & self control & interactive control\tabularnewline
\midrule 
1 & \ding{55} & \ding{55} & 42.0 & 65.1\tabularnewline

2 & \ding{51} & \ding{55} & 42.3 & 65.8\tabularnewline

3 & \ding{51} & \ding{51} & 42.4 & 65.8\tabularnewline

\bottomrule
\end{tabular}

  }
  {
  \captionof{table}{Evaluation of bidirectional bridging (BiB) module. $F_r\rightarrow F_c$ denotes that the control embedding $F_c$ is enhanced by information from the visual embedding $F_r$ and $F_c\rightarrow F_r$ denotes that the visual embedding $F_r$ is enhanced by the control embedding $F_c$.}
 \label{tab:ablation_beb}
  }
  \end{floatrow}
\end{table*}
\vspace{-2mm}
\section{Conclusion}
\label{sec:conclusion}

We proposed \Ours, a new region-level captioning paradigm with expanded capacity to overcome the caption degeneration issue by introducing control words.
\Ours consists of three components: visual embedding extraction, control embedding generation, and controllable caption generation.
The visual embedding extraction component can extract detail-rich and context-aware vision features. The control embedding generation component introduces a discriminative model to predict control words with less ambiguity, while the controllable caption generation component constrains \Ours to generate captions within a few sub-spaces containing the control words. In this way, \Ours increases the opportunity of hitting less frequent captions to alleviate the caption degeneration issue.
% %
During testing, when providing interactive control words from human or expert models, the model can generate captions beyond the caption space during training, demonstrating the model’s generalization ability.
\Ours sets a solid baseline for the challenging region-level captioning task and provides fresh insight about regularizing the caption space.

\clearpage
\bibliographystyle{splncs04}
\bibliography{main}
\end{document}